\documentclass[a4paper,twoside]{article}

\usepackage{epsfig}
\usepackage{subcaption}
\usepackage{calc}
\usepackage{amssymb}
\usepackage{amstext}
\usepackage{amsmath}
\usepackage{amsthm}
\usepackage{multicol}
\usepackage{pslatex}
\usepackage{apalike}
\usepackage[bottom]{footmisc}
\usepackage{SCITEPRESS}     

\usepackage{multirow}
\usepackage{xcolor}
\usepackage{url}

\begin{document}

\title{Benchmarking person re-identification datasets and approaches for practical real-world implementations}

\author{
\authorname{Jose Huaman,Felix O. Sumari, Luigy Machaca, Esteban Clua and Joris Guerin}
 \affiliation{\sup{1} Instituto de Computação, Universidade Federal Fluminense, Niteroi-RJ , Brazil}
 \affiliation{\sup{2}LAAS-CNRS, Université de Toulouse, Toulouse, France}
 \email{\{jmhcruz, fsumari, luigyarcana\}@id.uff.br,esteban@ic.uff.br, jorisguerin.research@gmail.com}
 }


\abstract{Recently, Person Re-Identification (Re-ID) has received a lot of attention. Large datasets containing labeled images of various individuals have been released, allowing researchers to develop and test many successful approaches. However, when such Re-ID models are deployed in new cities or environments, the task of searching for people within a network of security cameras is likely to face an important domain shift, thus resulting in decreased performance. Indeed, while most public datasets were collected in a limited geographic area, images from a new city present different features (e.g., people's ethnicity and clothing style, weather, architecture, etc.). In addition, the whole frames of the video streams must be converted into cropped images of people using pedestrian detection models, which behave differently from the human annotators who created the dataset used for training. To better understand the extent of this issue, this paper introduces a complete methodology to evaluate Re-ID approaches and training datasets with respect to their suitability for unsupervised deployment for live operations. This method is used to benchmark four Re-ID approaches on three datasets, providing insight and guidelines that can help to design better Re-ID pipelines in the future.}

\onecolumn \maketitle \normalsize \setcounter{footnote}{0} \vfill

This paper is the extended version of our short paper accepted in VISAPP - 2023.

\section{\uppercase{Introduction}}
\label{sec:introduction}


As many cameras are being deployed in public places (airports, malls, etc.), monitoring of video streams by security agents becomes impractical. Automated processing is a promising perspective to analyze the whole network in real-time, and select only relevant sequences for verification by human operators. This paper deals with person Re-Identification (Re-ID), a computer vision problem to find an individual in a network of non-overlapping cameras~\cite{old_reID_survey}. It has diverse potential applications such as suspect searching~\cite{open_set_first_paper}, identifying owners of abandoned luggage~\cite{abandoned_luggage}, or recovering missing children~\cite{missing_children}.
In the literature, the problem of Re-ID is studied under different settings (see 
Section~\ref{sec:related_paradigms}). On the one hand, the most studied Re-ID paradigm, which we refer to as \emph{standard Re-ID}, tries to find images representing the query person within a gallery of pre-cropped images of persons, containing at least one correct match~\cite{standard_reID_survey}. 
On the other hand, we recently introduced a setting considering specifically the constraints to implement Re-ID for live operations, which we call as \emph{live Re-ID}~\cite{application_level_ReID}. The first contribution of this paper is to formalize the definition and constraints associated with live Re-ID and to extend the evaluation metrics presented in~\cite{application_level_ReID} to facilitate interpretation.

\begin{figure*}[t]
    \centering
    \includegraphics[width=0.75\textwidth]{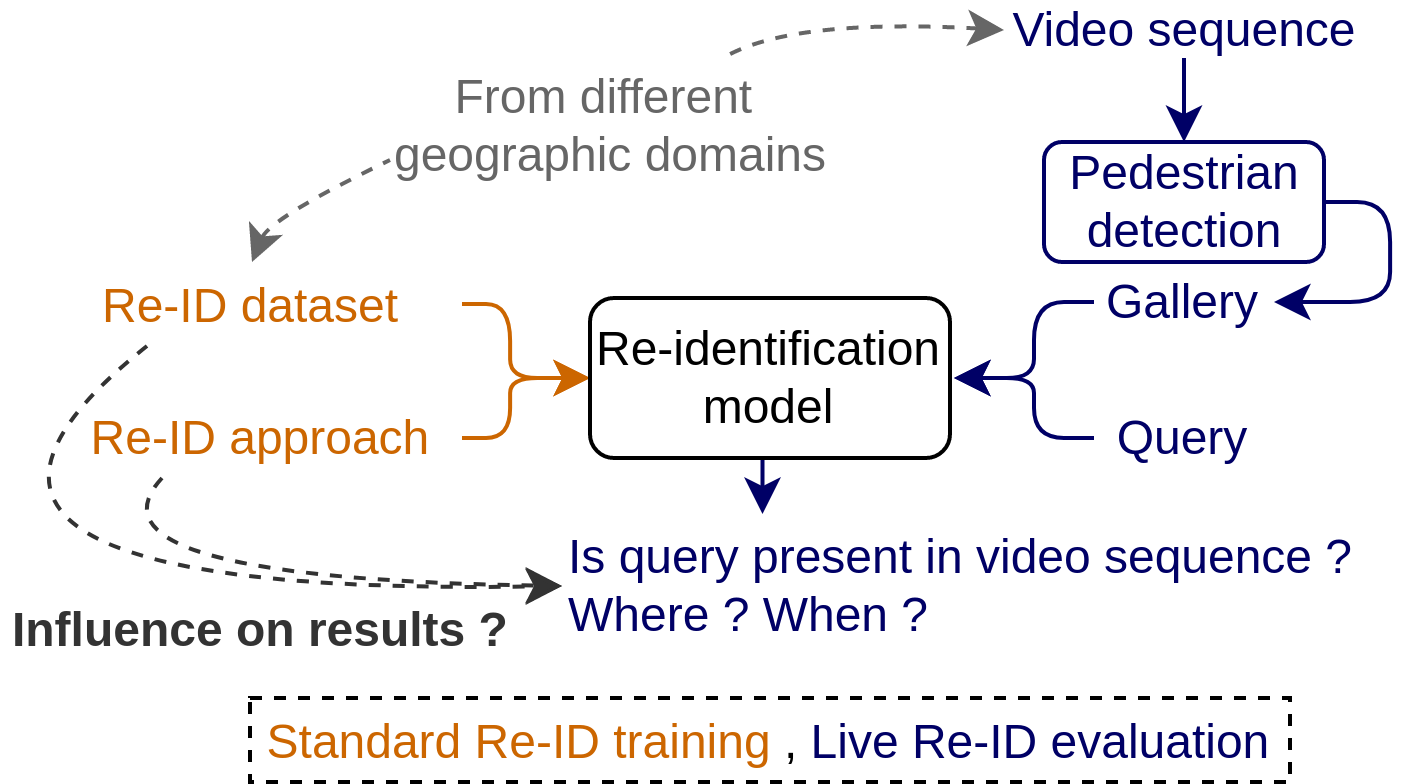}
    \caption{\textbf{Conceptual overview}. Representation of the objectives of our benchmark study. This work aims at evaluating how different standard Re-ID approaches and training datasets behave for practical deployment in new environments (live Re-ID).}
    \label{fig:overview}
\end{figure*}

Standard Re-ID is not the best-suited paradigm for practical implementations, as it does not consider the influence of domain shift due to pedestrian detection errors or deployment in a city with different characteristics. Indeed, in our previous experiments~\cite{application_level_ReID}, we showed that training a successful Re-ID model with respect to standard Re-ID metrics does not guarantee good performance when evaluated in a specific live Re-ID context. 
Nevertheless, most publicly available large-scale datasets for Re-ID focus on the standard Re-ID setting, and many successful approaches have been developed for this specific purpose. For this reason, we believe that it is essential to study if these datasets and approaches can be used to implement and deploy practical applications in different contexts. 
More specifically, the objective of this paper is to answer the following questions: \begin{enumerate}
    \item Which characteristics of a standard Re-ID dataset (diversity, size) are most important to train standard Re-ID models for the live Re-ID setting?
    \item Which standard Re-ID approaches can be successfully deployed for practical implementations in the live Re-ID setting? 
    \item Do different Re-ID approaches have different optimal datasets for deployment?
    \item Can we use a simple cross-dataset evaluation methodology to assess the deployability of a given approach-dataset pair?
\end{enumerate} 
To answer these questions, we conducted a study using three standard Re-ID training datasets and four recent standard Re-ID approaches. For each approach-dataset pair, the Re-ID model obtained was evaluated against the other two datasets and against another one configured for the live Re-ID setting. We also combine training datasets to investigate how dataset size and diversity influence the generalization of the obtained standard Re-ID model. A conceptual overview of our objectives is represented in Figure~\ref{fig:overview}.


In this paper, we consider the evaluation of Re-ID models without additional training on images from the target domain. More sophisticated approaches have been proposed for domain adaptation of standard Re-ID models. On the one hand, the unsupervised domain adaptation problem consists in leveraging unlabeled data from the target domain to improve the performance of the standard Re-ID model~\cite{zhao2020unsupervised,mekhazni2020unsupervised}. On the other, methods from the transfer learning field~\cite{zhao2020unsupervised} have been applied to fine-tune standard Re-ID models for new contexts where a small amount of labeled data is available~\cite{chen2018deep}. Such domain adaptation approaches are not tested in this work, but we believe standard Re-ID models performing well without target domain training (our experiments) are likely to be good initialization for more sophisticated fine-tuning approaches. On another note, it was shown that considering bounding box extraction and Re-ID separately is not as good as end-to-end approaches for person search, i.e., galleries of whole scene images~\cite{person_search_explain}. However, our results show that this two-step approach can perform well on the live Re-ID setting for some configurations. Likewise, we believe that the results from our study can be useful to pre-train successful initial live Re-ID models and to guide the development of more complex end-to-end architectures for live Re-ID.

This paper is organized as follows: Section~\ref{sec:related} discusses the relevant related literature. The proposed benchmark methodology is detailed in Section~\ref{sec:methodology}. The results are presented in Section~\ref{sec:results} and discussed in Section~\ref{sec:discussion}. Finally, Section~\ref{sec:conclusion} presents our conclusions.

\section{\uppercase{Related work}}
\label{sec:related}
A complete literature review of Re-ID approaches is not the purpose of this paper. Instead, we present clear definitions of the different existing Re-ID settings and discuss existing benchmark studies about Re-ID.

\subsection{Person re-identification settings}\label{sec:related_paradigms}
The field of Re-ID consists in retrieving instances of a given individual, called the \emph{query}, within a complex set of multimedia content called the \emph{gallery}~\cite{standard_reID_first_paper}. Different settings are defined by how they represent the query and the gallery items, the constraints on the gallery content, the boundaries of the Re-ID system, and the evaluation methodology.

\subsubsection{Popular settings}

\paragraph{Standard Re-ID} In this setting, both the query, and all items in the gallery are well-cropped images representing entire human bodies. It is sometimes called closed-set Re-ID as it assumes that the query has at least one representative in the gallery. According to the statistics in Papers with Code~\cite{papers_with_code} it is the most studied Re-ID setting by a large margin, in terms of number of papers, datasets and benchmarks published. Some standard Re-ID datasets and successful methods are used for our benchmark study and presented in Section~\ref{sec:methodology}. For a more complete overview of standard Re-ID approaches, see~\cite{standard_reID_survey2}.

\begin{figure*}[t]
    \centering
    \includegraphics[width=0.9\textwidth]{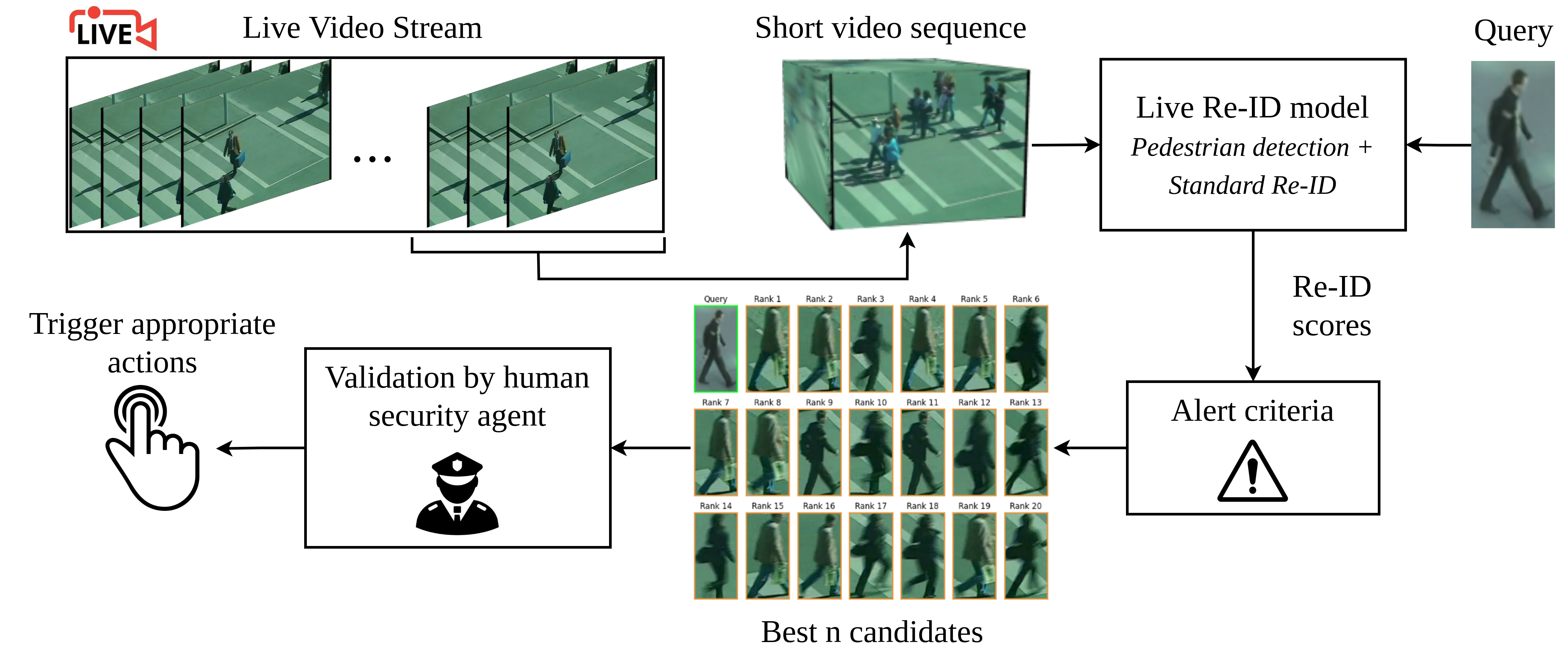}
    \caption{\textbf{The live Re-ID setting}. When deploying Re-ID models in practice, the galleries are composed of whole scene video sequences. When an alert is raised, the data are verified by a security agent to decide whether actions should be triggered.}
    \label{fig:liveReId}
\end{figure*}

\paragraph{Person search} This setting consists in replacing the gallery items by whole scene images~\cite{person_search_explain}. In other words, a person search model must return the index of the gallery image where the query is present and its location in terms of Bounding Box (BB) coordinates. A survey about person search approaches was proposed in~\cite{person_search_survey}.

\paragraph{Open-set Re-ID} This setting differs from standard Re-ID in that there is no guarantee that the query is represented in the gallery, i.e., an open-set Re-ID model should be able to answer whether the gallery contains the query. The reader can refer to~\cite{open_set_survey} for an overview of recent approaches.

\paragraph{Video-based Re-ID} In this setting, images (query and gallery) are replaced by image sequences extracted from consecutive video frames. Sequences are composed of well-cropped entire body images representing the same person. A complete survey of video-based Re-ID was proposed in~\cite{standard_reID_survey2}.

\paragraph{Others} For completeness, we mention other Re-ID variants, namely unsupervised Re-ID\cite{unsupervised_reid}, semi-supervised Re-ID\cite{semisupervised_reid}, human-in-the-loop Re-ID\cite{human_in_the_loop_reid}, or federated Re-ID\cite{federated_ReID}. However, their specificity lie in Re-ID models training while the other settings above focus on constraints at inference.

\subsubsection{Live Re-ID setting} \label{sec:related_live}

In this section, we clearly define and formalize the \emph{live Re-ID} setting, which is inspired by our previous work~\cite{application_level_ReID}. It takes into account all relevant aspects for deploying Re-ID in practical real-world applications. An overview of the live Re-ID workflow can be seen in Figure~\ref{fig:liveReId}.

When looking for a query person during live operations, whole scene videos need to be processed in near real-time, hence the galleries for live Re-ID are composed of the consecutive \emph{whole scene frames} from \emph{short video sequences}. The live Re-ID context is also highly \emph{open-set} as the probability to have the query in a short video sequence from a given camera is low. Hence, this setting combines elements from several of the Re-ID settings mentioned above. Using these live Re-ID characteristics, it was recently shown that using tracking and anomaly detection to reduce the size of the generated gallery improves live Re-ID results~\cite{machaca2022trade}. 

Another key characteristic of live Re-ID is that the training context is different from the deployment context. Indeed, building new specialized datasets for deployment in every shopping mall or small city is unrealistic from the perspective of future advances in the field. This highlights the importance of studying \emph{cross domain} transfer of Re-ID, which was first discussed and highlighted in~\cite{luo_strong_2020}.

Finally, this setting also takes into account that Re-ID model predictions need to be \emph{processed by a human agent}, who triggers appropriate actions. This way, very high rank-1 accuracy is not mandatory for live Re-ID, as the operator can find the query in later ranks. On the other hand, false alarm rates must be kept low to avoid overloading human operators. To evaluate these two objectives, we introduced two evaluation metrics representing both dimensions of the problem in~\cite{application_level_ReID}. The experiments conducted in this paper aim at studying the transferability of standard Re-ID approaches and datasets for deployment in the live Re-ID setting.



\subsection{Person re-identification benchmarks}\label{sec:related_eval_methods}

This section presents several benchmark studies considering different aspects of the Re-ID pipeline. 

A large scale benchmark experiment was conducted to compare various approaches for standard and video-based Re-ID~\cite{benchmark_reID_single_multi}. They evaluated more than 30 approaches on 16 public datasets, and produced the largest Re-ID benchmark to date. In addition, they built a new dataset to represent several constraints for real-world implementations, such as pedestrian detection errors and illumination variations, among others. However, they do not consider cross domain performance 
and all evaluations are conducted in the closed-set setting, which are major limitations regarding future deployments. 
In addition, a smaller systematic evaluation of video-based Re-ID approaches was proposed in~\cite{mars}.

Another extensive set of experiments was conducted to evaluate different pedestrian detection models on a two-step person search pipeline~\cite{benchmark_OD_reID}. They demonstrated that the best models on standard object detection metrics are not necessarily the best suited for Re-ID from whole scene frames. In addition, a first benchmark regarding cross-domain transfer of Re-ID approaches was proposed in~\cite{benchmark_reID_cross}. Their experiments consisted in training an approach on one standard Re-ID dataset and evaluating on another. Finally, on another note, different approaches for federated Re-ID where compared in~\cite{federated_ReID}.

The studies presented above have brought valuable insights to the Re-ID community. However, none of them allows to assess the performance of a Re-ID model against all the challenges involved during deployment in a new environment for practical use in security applications. Our paper contributes to bridging this gap by conducting experiments within the live Re-ID setting, which was designed to take into account all these challenges. In particular, we consider the influence of different standard Re-ID approaches and training datasets on live Re-ID results.

\section{\uppercase{Benchmark methodology}}\label{sec:methodology}
The objective of this paper is to study if different standard Re-ID approaches and training datasets can be used to build efficient live Re-ID pipelines, ready for practical deployment. 
This section presents the different components of the proposed benchmarking evaluation, i.e., the compared datasets and approaches, metrics used, and experiments conducted.

\begin{figure}[t]
    \centering
        \begin{subfigure}{0.48\textwidth}
        \centering
        \includegraphics[width=0.95\textwidth]{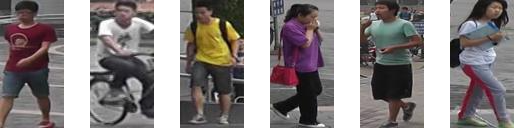}
        \caption{Market-1501.}
        \label{fig:datasets_market}
        \end{subfigure}
        ~
        
        \begin{subfigure}{0.48\textwidth}
        \centering
        \includegraphics[width=0.95\textwidth]{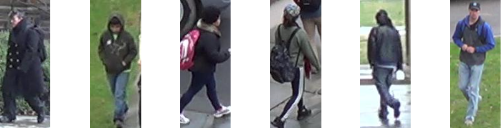}
        \caption{DukeMTMC.}
        \label{fig:datasets_duke}
        \end{subfigure}
        ~
        
        \begin{subfigure}{0.48\textwidth}
        \centering
        \includegraphics[width=0.95\textwidth]{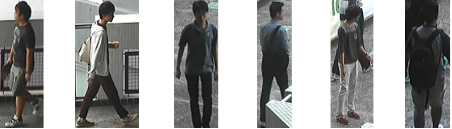}
        \caption{CUHK03.}
        \label{fig:datasets_cuhk}
        \end{subfigure}
        ~
        
        \begin{subfigure}{0.48\textwidth}
        \centering
        \includegraphics[width=0.95\textwidth]{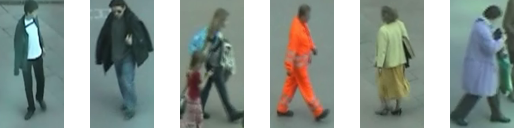}
        \caption{Original PRID-2011.}
        \label{fig:datasets_prid}
        \end{subfigure}
        ~
        
        \begin{subfigure}{0.48\textwidth}
        \centering
        \includegraphics[width=0.95\textwidth]{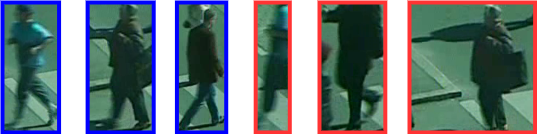}
        \caption{{m-PRID}. Extracted from PRID-2011 videos using YOLO-V3. Blue indicates good images for standard Re-ID, while red BB are likely to generate Re-ID errors.}
        \label{fig:datasets_mprid}
        \end{subfigure}
        ~
        
    \caption{\textbf{Benchmarking datasets}. Example images from the datasets used in our experimental study.}
    \label{fig:repredata}
\end{figure}

\subsection{Datasets}
In our experiments, we used three public datasets to train and evaluate standard Re-ID models, and a live Re-ID dataset to evaluate the trained Re-ID models within the context of live operations. Figure~\ref{fig:repredata} shows example images from the datasets, where we can see that they represent people from different geographic regions, under different resolutions, lighting conditions, and camera angles.

\subsubsection{Standard Re-ID datasets}

\begin{table*}[t]
\centering
\caption{\textbf{Standard Re-ID training datasets}. Characteristics of the standard Re-ID datasets used in this paper.}
\label{tab:data_table}
\begin{center}
\begin{tabular}{cccccc}
\cline{1-6}
Dataset &
\# Cameras &
Split &
Input type & 
\# IDs & \# Images \\
\cline{1-6}
\multirow{3}{*}{CUHK03} & \multirow{3}{*}{2} & 
Train & -- & 767 & 7368 \\
& & \multirow{2}{*}{Test} & Query & 700 & 1400 \\ 
& & & Gallery & 700 & 5328 \\
\cline{1-6}
\multirow{3}{*}{DukeMTMC} & \multirow{3}{*}{8} & 
Train & -- & 702 & 16522 \\ 
& & \multirow{2}{*}{Test} & Query & 702 & 2228 \\ 
& & & Gallery & 1110 & 17661 \\ 
\cline{1-6}
\multirow{3}{*}{Market-1501} & \multirow{3}{*}{6} & 
Train & -- & 751 & 12936\\
& & \multirow{2}{*}{Test} & Query & 750 & 3368 \\
& & & Gallery & 751 & 15913 \\
\cline{1-6}
\end{tabular}
\end{center}
\end{table*}

This section presents briefly the standard Re-ID datasets used in this study. Table~\ref{tab:data_table} summarizes relevant statistics.


\paragraph{Market-1501} This dataset was collected at a supermarket in Tsinghua University, Beijing, China~\cite{market}.  
The cropped images are detected automatically using a Deformable Part Model~
\cite{DPM}, which outputs are filtered manually to keep only good BB representing human bodies. This automated way of extracting BB is closer to realistic settings, which might improve live Re-ID results for models trained on Market-1501. As illustrated by Figure~\ref{fig:datasets_market}, cropped images appear to present a high level of details about the people represented (i.e., images are taken from a close perspective or videos are high resolutions). Lighting conditions in this dataset are also good to distinguish specific features.

\paragraph{DukeMTMC} 
This dataset was collected at the Duke University campus, Durham, North Carolina, USA~\cite{duke}. 
The BB in DukeMTMC are hand drawn, lighting conditions are good but resolution of the BB is relatively low (Figure~\ref{fig:datasets_duke}). 

\paragraph{CUHK03} 
This dataset was built using video footage collected at the campus of the Chinese University of Hong Kong~\cite{cuhk}. In our work, we used the manually labeled version of the BB. Cropped images are high resolution but illumination is dark, which reduces images quality (Figure~\ref{fig:datasets_cuhk}). 

\subsubsection{Live Re-ID dataset}
To evaluate the different standard Re-ID models for the live Re-ID setting, we used the same dataset as our previous work~\cite{application_level_ReID}, which we call \emph{m-PRID}. It is a modified version of PRID-2011~\cite{prid2011}, built from the raw video footage and the original annotations that were used to create the official curated version of PRID-2011\footnote{We thank the authors of PRID-2011 for their help.}. The videos were collected from two non-overlapping cameras (A and B), located in Graz, Austria. This way, compared to the training datasets above, evaluation on m-PRID represents a geographic domain shift. In total PRID-2011 contains 385 different identities for camera A and 749 for camera B, of which 200 identities appear in both cameras. The m-PRID dataset is composed of several two minutes videos (30 from A and 33 from B). For each short video sample, a ground truth file gather information about each person it contains (identifier, frames where it appears, bounding box coordinates). For evaluation, a total of 73 queries are considered.

To better grasp the influence of the pedestrian detection model, we also evaluate our different models on the original PRID-2011 dataset for standard Re-ID. Figure~\ref{fig:datasets_prid} shows cropped images of poor resolution, taken from relatively high camera angle compared to other datasets. This way, we can see if performance decrease on the live Re-ID setting are due to the domain shift of PRID or to the pedestrian detector inaccuracies (Figure~\ref{fig:datasets_mprid}). 

\subsection{Re-ID approaches evaluated} \label{sec:reid_approaches}
This work studies the performance of four successful standard Re-ID approaches. To complement previous benchmark studies (Section~\ref{sec:related_eval_methods}), only recent approaches are selected for this work. 

\subsubsection{Bag of Tricks (BoT)} The \emph{Bag of Tricks} approach resulted from the observation that most improvements for Re-ID baselines come from neural network training tricks rather than Re-ID approaches themselves~\cite{luo_bag_2019}. As a result, they came up with a simple recipe to successfully train standard Re-ID models on top of a ResNet-50 backbone~\cite{he_deep_2015}. In particular: 
\begin{enumerate}
    \item the model is pretrained on ImageNet,  
    \item the dimension of the fully connected layer is set to the number of training identities, 
    \item the batch size is set to 64, 
    \item images are resized to 256 X 128, 
    \item both 
    triplet and cross entropy loss are used, and 
    \item Adam is adopted for optimization.
\end{enumerate}


\subsubsection{Strong Baseline and Batch Normalization Neck (SBS)} This approach~\cite{luo_strong_2020} extended BoT by adding the following tricks: 
\begin{enumerate}
    \item a warm-up strategy~\cite{Fan_2019} is applied to bootstrap the network, 
    \item random erasing augmentation~\cite{zhong_random_2017} is used to account for potential occlusion, 
    \item label smoothing~\cite{szegedy_rethinking_2015} is used to reduce overfitting,
    \item the last stride of the ResNet-50 backbone is set to 1 to increase spatial resolution, 
    \item batch normalization layers are added, and
    \item a new center loss is introduced to account for the clustering effect of tracking.
\end{enumerate}


\subsubsection{Attention Generalized mean pooling with Weighted triplet loss (AGW)} This technique~\cite{standard_reID_survey2} was also designed on top of BoT with three major improved components:
\begin{enumerate}
    \item 
    a powerful non-local attention block is developed to mix the part and global attention features,
    \item a learnable pooling layer replaces max and average pooling to better capture the domain-specific discriminative features, 
    \item the use of weighted regularization triplet loss inherits the advantages of relative distance optimization between positive and negative pairs without introducing additional parameters.
\end{enumerate}

\subsubsection{Multiple Granularity Network (MGN)} This approach~\cite{wang_learning_2018} was designed
combines local and global information in different image granularity. 
Its specificities are:
\begin{enumerate}
    \item a strategy to learn local feature representations by splitting the input image in horizontal stripes. Then, the ResNet-50 backbone is divided into a multiple branches after the fourth residual stage.
    \item a strategy to learn global features using a global branch using down-sampling with a stride-2 convolution layer and representations with 256 dimension features. The part-N branch has a similar architecture without down-sampling.
    \item a new Loss functions, combining Softmax for classification and tripĺet loss for metric learning.
\end{enumerate}

\subsection{Proposed experiments}
To compare the Re-ID datasets and approaches presented above, several experiments are conducted.

\subsubsection{Single dataset evaluation}
We first evaluate each approach and dataset pair individually. The standard Re-ID approach is simply fitted to the training split of the dataset, and evaluated on the testing split. The quality of the Re-ID model's predictions on the testing set is assessed using standard Re-ID metrics, coming from the field of information retrieval:

\paragraph{Rank-n} 
This metric represents the proportion of queries for which at least one correct match was predicted within the $n$ highest ranked gallery images~\cite{cmc_curve}. In practice, we report results for $n\in\{1,5,10\}$. This metric represents the Re-ID model's ability to retrieve the easiest match.

\paragraph{mAP} The \emph{mean average precision} for Re-ID takes into account the predicted ranks of all existing matches~\cite{benchmark1}. To have a perfect mAP, all the gallery images corresponding to the query need to be ranked in the first places. It is the average performance across all instances of the query.

\paragraph{mINP} The \emph{mean inverse negative penalty} reflects the position of the worst ranked match from the gallery~\cite{standard_reID_survey2}. It represents the capacity of a Re-ID model to find all instances of the query in the gallery.

These three metrics represent different skills of a Re-ID model. Computing them might help understand which of these skill is important regarding generalization to new contexts and to more complex real-world scenarios, i.e., live Re-ID in different cities.

\subsubsection{Cross-dataset evaluation}
A simple cross-dataset experiment is also conducted~\cite{benchmark_reID_cross}. It consists in training an approach on one of the three standard Re-ID datasets, and evaluating it on the other two. The same metrics are used (rank-n, mAP and mINP). As the datasets were built in different geographic areas, these results can give first insights about domain generalization of the different training datasets and approaches. Conducting such cross-dataset evaluation is also much easier than evaluating the system in the live Re-ID setting. Hence, another objective of this experiment is to discover if simple cross-dataset evaluation can be used as a proxy to quickly test new datasets and approaches for live implementations. In other words, we want to know if there is correlation between cross-dataset results and live Re-ID results of dataset-approach pairs.

For the cross-datasets experiments, we also try to combine training datasets to see if it improves test performance. In the COMBINED$_{\text{all}}$ experiments, training is conducted on all training sets available (Market-1501, DukeMTMC and CUHK03), including the one corresponding to the test set of interest. This allows to evaluate if adding data from other sources can help improving standard Re-ID in the traditional supervised setting. In the COMBINED$_{\text{others}}$ experiments, the training set corresponding to the test dataset is excluded. For example, when evaluating on CUHK03, the standard Re-ID models are trained on Market-1501 and DukeMTMC. Finally, the COMBINED$_{\text{scaled}}$ setting is similar to COMBINED$_{\text{others}}$, but we ensure that the total number of training data is equal to the number of data in the largest dataset. 

For example, when combining datasets $A$ and $B$, respectively of size $N_A$ and $N_B$, we only take fractions $N^*_A$ and $N^*_B$ of each datasets such that $N^*_A + N^*_B = \text{max}(N_A, N_B)$ and $N^*_A = N^*_B$. Comparing COMBINED$_{\text{scaled}}$, with COMBINED$_{\text{others}}$ allows to evaluate how size and diversity affect the generalization power of a dataset. For evaluations on PRID-2011, which is not among the training datasets, COMBINED$_{\text{all}}$ and COMBINED$_{\text{others}}$ are identical and referred to as COMBINED. When evaluating on CUHK03, the COMBINED$_{\text{scaled}}$ dataset is composed of 8261 images from DukeMTMC and 8261 from Market-1501. When evaluating on DukeMTMC it contains 6468 images from both CUHK03 and Market-1501, and when evaluating on Market-1501 it contains 9754 images from DukeMTMC and 7368 images from CUHK03. Finally, when evaluating on PRID-2011, COMBINED$_{\text{scaled}}$ is composed of 5507 images from CUHK03, 5508 from DukeMTMC and 5507 from Market-1501.
    
\subsubsection{Live Re-ID evaluation}\label{sec:methodology_livereID_eval}

Finally, each standard Re-ID approach and dataset pairs are evaluated in the live Re-ID setting using the m-PRID dataset. We apply the evaluation methodology from~\cite{application_level_ReID}. For each short video sequence, Bounding Boxes (BB) of pedestrians are extracted using a YOLO-V3 object detector~\cite{redmon2018yolov3}, trained on COCO~\cite{lin2015microsoft} and available in TensorFlow~\cite{tensorflow2015-whitepaper}. The score threshold used to decide which predicted BB to keep is set to $0.5$. Then, the trained standard Re-ID approaches are applied to the gallery composed of these BB. Following the notations of~\cite{application_level_ReID}, the length of video sequences evaluated $\tau$ is set to $1000$ frames and the number of candidates shown to the monitoring agent $\eta$ is set to $20$. These values generated best results by a large margin in their experiments. For the threshold $\beta$ on Re-ID scores used to generate alerts, we test all values between $0$ and $1$ with a step size of $0.02$.

To compare the different models, we use the live Re-ID metrics introduced in~\cite{application_level_ReID}. On the one hand, the \emph{Finding Rate} (FR) represents the proportion of videos where the query was present, such that an alert was shown to the agent and where the query was among the selected candidates. A low FR means that the query was missed frequently. On the other, the \emph{True Validation Rate} (TVR) represents the proportion of alert shown to the monitoring agent in which the query was present among the candidates. A low TVR means that the agent was frequently disturbed for no reason, which can be problematic when many cameras need to be monitored simultaneously. 

In this paper we also propose two new metrics to represent the performance of a live Re-ID approach with a single number, to facilitate comparisons and interpretation. The first one is based on the observation that 
the meanings of FR and TVR are respectively very close to recall and precision. This way, similarly to object detection evaluation, we can plot \emph{TVR vs FR} curves and compute the \emph{mean Average Precision} ({mAP}) as the area under the curve. The second unified metric 
consists in computing a weighted harmonic mean of FR and TVR, similarly to the F-score computation for precision and recall. We call the resulting metric $F_{\gamma}$, which is defined as follows:
\begin{equation}
    F_{\gamma} = (1+\gamma^2).\frac{\text{FR}.\text{TVR}}{(\gamma^2.\text{FR})+\text{TVR}}.
\end{equation}

In practice, we compute $F_{\gamma}$ for $\gamma \in \{0.5, 1, 2\}$. In $F_{0.5}$, we consider that having a high TVR is two times more important than a high FR. In $F_{2}$, we consider FR two times more important than TVR, and in $F_{1}$ FR and TVR contribute equally to the results. However, for each value of the threshold $\beta$, there is a different corresponding value of $F_{\gamma}$. To solve this issue, we use the same approach as Guerin et al.,\cite{icmla_joris} consisting in evaluating a model by its performance at the optimal configuration. The result is called \emph{optimal $F_{\gamma}$} ($F^*_{\gamma}$), and corresponds to the highest $F_{\gamma}$ across values of $\beta$. The value of $\beta$ corresponding to $F^*_{\gamma}$ can be viewed as the operating point of the Re-ID model, which can be obtain by quick experiments in the practical implementation context. An $F^*_{\gamma}$ score of 1 means that there exist a Re-ID threshold $\beta$ such that it always find the query when it is in the video sequence, but never raises alerts when it is not.

\begin{table*}[t]
\caption{\textbf{Single dataset evaluations}. Results obtained by training and evaluating Re-ID approaches with the train and test splits of the same dataset. For each dataset, the best Re-ID approach is in bold.}
\label{tab:classic_reid_eval}
\begin{center}
\scalebox{0.9}{
\begin{tabular}{ccccccc}%
\hline
Dataset & Approach & Rank-1 & Rank-5 & Rank-10 & mAP & mINP \\ 
\hline
\multirow{4}{*}{CUHK03} & AGW & 0.73 & 0.88 & 0.92 & 0.72 & 0.63 \\
 & MGN & \textbf{0.78} & \textbf{0.91} & \textbf{0.95} & \textbf{0.76} & \textbf{0.66} \\
 & SBS & 0.74 & 0.89 & 0.93 & 0.73 & 0.62 \\
 & BoT & 0.69 & 0.86 & 0.92 & 0.67 & 0.55 \\ \hline
\multirow{4}{*}{DukeMTMC} & AGW & 0.89 & 0.95 & \textbf{0.97} & 0.80 & 0.46 \\
 & MGN & \textbf{0.91} & \textbf{0.96} & 0.97 & \textbf{0.82} & \textbf{0.47} \\
 & SBS & 0.89 & 0.95 & 0.96 & 0.79 & 0.44 \\
 & BoT & 0.87 & 0.94 & 0.96 & 0.77 & 0.41 \\ \hline
\multirow{4}{*}{Market-1501} & AGW & 0.95 & 0.99 & 0.99 & 0.88 & 0.66 \\
 & MGN & \textbf{0.96} & \textbf{0.99} & \textbf{0.99} & \textbf{0.89} & \textbf{0.66} \\
 & SBS & 0.95 & 0.98 & 0.99 & 0.88 & 0.66 \\
 & BoT & 0.94 & 0.98 & 0.99 & 0.86 & 0.61 \\ 
\hline
\end{tabular}
}
\end{center}
\end{table*}

The objective of this experiments is to see if the best approaches and datasets from previous experiments are also the best ones from the perspective of practical implementation in new cities.

Combined datasets experiments are also conducted for the live Re-ID setting. However, as PRID-2011 is not one of the training datasets used in our experiments, COMBINED$_{\text{all}}$ and COMBINED$_{\text{others}}$ are actually equivalent here and simply referred to as COMBINED. They are also compared against COMBINED$_{\text{scaled}}$ results to study the impact of dataset size and diversity. The COMBINED$_{\text{scaled}}$ training dataset for live Re-ID experiments on m-PRID are composed of 5507 images from CUHK03, 5508 from DukeMTMC and 5507 from Market-1501.

    
    
    

\section{\uppercase{Results}} \label{sec:results}


In order to improve clarity, only a condensed version of the results is presented. The complete results can be found at: \url{https://github.com/josemiki/benchmarking_person_Re_ID}. It contains all the results from cross-dataset evaluation, contains the missing metrics and the TVR vs FR curves for live Re-ID evaluations. Overall, the curated results presented in the core paper are representative of the complete results and are sufficient to draw our conclusions.

The results for single dataset evaluation are reported in Table~\ref{tab:classic_reid_eval}. The results obtained are good: rank-1 and mAP are around 70\% for the worst approach on the most difficult dataset. They are also relatively homogeneous: for each dataset-metric pairs, all four methods perform similarly (less than 10\% difference). The results obtained show that the tested approaches generalize differently to new contexts. For instance training MGN on Market-1501 leads to 47\% rank-10 accuracy on CUHK03, while the same experiment using BoT only reaches 15\%. For comparison, when training was conducted on CUHK03 itself, only a 3\% difference was observed between the approaches (Table~\ref{tab:classic_reid_eval}). The choice of the training dataset is also important, e.g., when training MGN for CUHK03, Market-1501 is 13\% better than DukeMTMC.

\begin{table*}[t]
\caption{\textbf{Cross-dataset evaluations}. Results obtained by training Re-ID approaches on one dataset and evaluating on another. For each evaluation dataset: the best Re-ID approach for a given dataset is in bold; the best training dataset for a given approach is in blue. R10 means Rank-10.}
\label{tab:cross_eval}
\begin{center}
\scalebox{0.9}{
\begin{tabular}{cccc|cc|cc|cc}
\hline
\multirow{2}{*}{\begin{tabular}[c]{@{}c@{}}Evaluation \\ dataset\end{tabular}} & \multirow{2}{*}{\begin{tabular}[c]{@{}c@{}}Training \\ dataset\end{tabular}} & \multicolumn{2}{c}{AGW} & \multicolumn{2}{c}{MGN} & \multicolumn{2}{c}{SBS} & \multicolumn{2}{c}{BoT} \\
& & R10 & mAP & R10 & mAP & R10 & mAP & R10 & mAP \\  
\hline
\multirow{5}{*}{CUHK03} & Market-1501 & 0.21 & 0.08 & \textbf{0.47} & \textbf{0.22} & 0.40 & 0.18 & 0.15 & 0.04 \\
& DukeMTMC & 0.18 & 0.06 & 0.34 & \textbf{0.14} & \textbf{0.35} & 0.13 & 0.15 & 0.05 \\
 & COMBINED$_{\text{all}}$ & \textcolor{blue}{0.94} & \textcolor{blue}{0.71} & \textcolor{blue}{\textbf{0.96}} & \textcolor{blue}{\textbf{0.82}} & \textcolor{blue}{0.94} & \textcolor{blue}{0.76} & \textcolor{blue}{0.92} & \textcolor{blue}{0.68} \\
 & COMBINED$_{\text{others}}$ & 0.32 & 0.14 & \textbf{0.55} & \textbf{0.27} & 0.52 & 0.24 & 0.28 & 0.11 \\
 & COMBINED$_{\text{scaled}}$ & 0.31 & 0.13 & \textbf{0.52} & \textbf{0.23} & 0.46 & 0.20 & 0.23 & 0.09 \\
\hline

\multirow{5}{*}{DukeMTMC} & Market-1501 & 0.58 & 0.22 & \textbf{0.77} & \textbf{0.39} & 0.74 & 0.34 & 0.49 & 0.15 \\
& CUHK03 & 0.50 & 0.17 & \textbf{0.70} & \textbf{0.31} & 0.60 & 0.21 & 0.36 & 0.10 \\ 
& COMBINED$_{\text{all}}$ & \textcolor{blue}{0.96} & \textcolor{blue}{0.79} & \textcolor{blue}{\textbf{0.97}} & \textcolor{blue}{\textbf{0.82}} & \textcolor{blue}{0.96} & \textcolor{blue}{0.78} & \textcolor{blue}{0.96} & \textcolor{blue}{0.77}\\
& COMBINED$_{\text{others}}$ & 0.65 & 0.29 & \textbf{0.81} & \textbf{0.44} & 0.79 & 0.41 & 0.55 & 0.21 \\ 
 & COMBINED$_{\text{scaled}}$ & 0.62 & 0.26 & \textbf{0.78} & \textbf{0.40} & 0.75 & 0.35 & 0.51 & 0.18 \\
\hline

\multirow{5}{*}{Market-1501} & DukeMTMC & 0.75 & 0.26 & \textbf{0.87} & \textbf{0.37} & 0.82 & 0.31 & 0.71 & 0.22 \\
& CUHK03 & 0.73 & 0.29 & \textbf{0.86} & \textbf{0.39} & 0.80 & 0.34 & 0.66 & 0.22 \\
& COMBINED$_{\text{all}}$ & \textcolor{blue}{0.99} & \textcolor{blue}{0.88} & \textcolor{blue}{\textbf{0.99}}& \textcolor{blue}{\textbf{0.91}} & \textcolor{blue}{0.99} & \textcolor{blue}{0.88} & \textcolor{blue}{0.99}& \textcolor{blue}{0.86} \\
& COMBINED$_{\text{others}}$ & 0.83 & 0.38 & \textbf{0.93} & \textbf{0.52}
& 0.91 & 0.47 & 0.80 & 0.34\\
 & COMBINED$_{\text{scaled}}$ & 0.83 & 0.38 & \textbf{0.92} & \textbf{0.52} & 0.89 & 0.46 & 0.78 & 0.32 \\
\hline

\multirow{5}{*}{PRID-2011} & CUHK03 & 0.18 & 0.11 & \textbf{0.35} & \textbf{0.26} & 0.29 & 0.20 & 0.13 & 0.09 \\
& DukeMTMC & 0.20 & 0.12 & \textbf{0.42} & \textbf{0.30} & 0.26 & 0.17 & 0.16 & 0.07 \\
& Market-1501 & 0.26 & 0.19 & \textbf{0.40} & \textbf{0.28} & 0.30 & 0.20 & 0.23 & 0.13 \\
& COMBINED & \textcolor{blue}{0.32} & \textcolor{blue}{0.20} & \textbf{0.45} & \textbf{0.35} & 0.33 & 0.23 & \textcolor{blue}{0.24} & \textcolor{blue}{0.15} \\
& COMBINED$_{\text{scaled}}$ & 0.24 & 0.18 & \textcolor{blue}{\textbf{0.46}} & \textcolor{blue}{\textbf{0.36}} & \textcolor{blue}{0.36} & \textcolor{blue}{0.26} & 0.22 & 0.15 \\

\hline
\end{tabular}
}
\end{center}
\end{table*}

The cross-dataset evaluation results are presented in Table~\ref{tab:cross_eval}. We only report Rank-10 for two reasons. First, the complete results show that the ranking of approaches is stable under different values of $n$. Second, having a high Rank-10 is more important than lower ranks for live Re-ID, as explained in Section~\ref{sec:related_live}.

Finally, the live Re-ID evaluation results are presented in Table~\ref{tab:live-reid}. They also illustrate that it is crucial to properly select the training dataset and approach for such task transfer. Overall, MGN appears to generalize much better for use in a live Re-ID setting. For training, Market-1501 appear to work best for most approaches except MGN. The best combination using a single dataset is MGN trained on DukeMTMC, reaching a mAP of 0.72 and an optimal F1 of 0.76.

\section{\uppercase{Discussion}} \label{sec:discussion}

In this section, we propose to discuss the results obtained in our study, highlighting several key insights regarding training standard Re-ID models for cross-domain live Re-ID. 

\subsection{Impact of the training dataset}

Proper selection of the training dataset clearly influences the results in a different evaluation domain. However, there is no clear winner between Market-1501 and DukeMTMC to know which individual dataset should be used for any context. In addition, the cross-dataset results do not allow to choose the best individual dataset for training models for the live Re-ID setting. Indeed, Table~\ref{tab:cross_eval} suggested that the best dataset for MGN should be Market-1501, whereas it is outperformed by DukeMTMC for live Re-ID (Table~\ref{tab:live-reid}). In the remaining of this section, we discuss the results obtained on the combined datasets settings to gain new insights regarding building standard Re-ID datasets for efficient training of live Re-ID models.

\subsubsection{Can data from a different domain improve results in the standard Re-ID scenario ?}

To answer this question, we compare results from Table~\ref{tab:classic_reid_eval} and the COMBINED$_{\text{all}}$ rows in Table~\ref{tab:cross_eval}. Overall, for both Rank-10 and mAP, the results for COMBINED$_{\text{all}}$ appear slightly better than the results obtained when learning only on the training set of the evaluated dataset. To illustrate this, we computed the mean and standard deviation across all evaluation dataset and approaches. When using only the training set we obtain the following results: $R10 = 0.962 \pm 0.026$ and $mAP = 0.798 \pm 0.068$. When combining the three available datasets for training we have: $R10 = 0.964 \pm 0.022$ and $mAP = 0.805 \pm 0.067$.

\begin{table*}[t]
\begin{center}
\caption{\textbf{Live Re-ID evaluation}. Results obtained by training Re-ID approaches on one standard Re-ID dataset and evaluating on m-PRID for the live Re-ID setting. For each training dataset, the best approach is in bold and for each approach, the best dataset is in blue.}
\label{tab:live-reid}
\scalebox{0.9}{
\begin{tabular}{ccc|cc|cc|cc|cc}
\hline
\multirow{2}{*}{Approach}& \multicolumn{2}{c}{CUHK03} & \multicolumn{2}{c}{DukeMTMC} & \multicolumn{2}{c}{Market-1501} & \multicolumn{2}{c}{COMBINED} & \multicolumn{2}{c}{COMBINED$_{\text{scaled}}$} \\
& $F_{\text{1}}^*$ & mAP & $F_{\text{1}}^*$ & mAP & $F_{\text{1}}^*$ & mAP & $F_{\text{1}}^*$ & mAP & $F_{\text{1}}^*$ & mAP\\ 
\hline
AGW & 0.39 & 0.23 & 0.40 & 0.25 &  0.46 & 0.33 & \textcolor{blue}{0.56} & \textcolor{blue}{0.49} & 0.49 & 0.39 \\
BoT & 0.27 & 0.10 & 0.40 & 0.22 & \textcolor{blue}{0.47} & \textcolor{blue}{0.32} & 0.45 & 0.30 & 0.44 & 0.31\\
SBS & 0.51 & 0.43 & 0.58 & 0.54 & 0.60 & 0.50 & \textcolor{blue}{0.71} & \textcolor{blue}{0.71} & 0.68 & 0.72\\
MGN & \textbf{0.66} & \textbf{0.60} & \textbf{0.76} & \textbf{0.72} & \textbf{0.69} & \textbf{0.63} & \textcolor{blue}{\textbf{0.81}} & \textcolor{blue}{\textbf{0.80}} & \textbf{0.77} & \textbf{0.75}\\ 
\hline
\end{tabular}
}
\end{center}
\end{table*}

To confirm this intuition, we conduct a Paired Sample T-Test to determine whether the mean difference between the results obtained using the single in-domain training set and the COMBINED$_{\text{all}}$ are statistically significant. The p-values obtained are $0.2750$ for R10 and $0.2313$ for mAP, suggesting that the Null Hypothesis cannot be rejected, i.e., \emph{we cannot conclude that using more data from a different domain is beneficial to the standard Re-ID training process}. 
    

\subsubsection{Between dataset size and diversity, which is most important for cross-domain generalization ?}

The first question we want to answer is whether combining datasets from different domains can help cross domain generalization. To evaluate this, we can compare the results for COMBINED$_{\text{others}}$ (COMBINED for PRID-2011) against the results from the best individual dataset in Table~\ref{tab:cross_eval}. The mean and standard deviation across all evaluation dataset and approaches are $R10 = 0.509 \pm 0.242$ and $mAP = 0.224 \pm 0.099$ for the best individual dataset, and $R10 = 0.580 \pm 0.241$ and $mAP = 0.297 \pm 0.124$ when combining all the available training datasets (except the one corresponding to the evaluated test set). The Paired Sample T-Test gives p-values of $0.0001$ for both R10 and mAP, which is extremely statistically significant. In other words, our experiments confirm that \emph{combining several training datasets from different domains allows to train Re-ID models that generalize better to new unknown domains}.

We then want to know if simply increasing the diversity in the training dataset without increasing its size also help for cross domain generalization. To evaluate this, we can compare the results for COMBINED$_{\text{scaled}}$ against the results from the best individual dataset in Table~\ref{tab:cross_eval}. As a reminder, COMBINED$_{\text{scaled}}$ consists in building a training dataset by taking data from all available training sets (except the one corresponding to the evaluated test set) in such a way that the total number of training data does not exceed that size of the largest individual training set. The mean and standard deviation across all evaluation dataset and approaches are $R10 = 0.509 \pm 0.242$ and $mAP = 0.224 \pm 0.099$ for the best individual dataset, and $R10 = 0.555 \pm 0.245$ and $mAP = 0.279 \pm 0.125$ for COMBINED$_{\text{scaled}}$. The Paired Sample T-Test gives p-values of $0.0001$ for R10 and $0.0005$ for mAP, which is extremely statistically significant. In other words, our experiments confirm that \emph{increasing diversity in training dataset, even without increasing its size, allows to train Re-ID models that generalize better to new domains}.

In view of the two encouraging results presented above, we now want to know whether the size of the training dataset is actually helping cross-domain generalization or if adding diversity is actually sufficient. To evaluate this, we can compare the results for COMBINED$_{\text{others}}$ against the results for COMBINED$_{\text{scaled}}$ in Table~\ref{tab:cross_eval}. The mean and standard deviation across all evaluation dataset and approaches are $R10 = 0.580 \pm 0.241$ and $mAP = 0.297 \pm 0.124$ for COMBINED$_{\text{others}}$, and $R10 = 0.555 \pm 0.245$ and $mAP = 0.279 \pm 0.125$ for COMBINED$_{\text{scaled}}$. The Paired Sample T-Test gives p-values of $0.0020$ for R10 and $0.0075$ for mAP, which is very statistically significant. In other words, our experiments confirm that \emph{adding more data from domains that are already present in the training set help generalization to new unknown domains}.

\subsubsection{live Re-ID results}

The live ReID results on m-PRID (Table~\ref{tab:live-reid}) confirm the conclusions drawn from the cross-dataset experiments. In particular, the COMBINED$_{\text{scaled}}$ results appear better than the results with a single training set, suggesting the importance of training data diversity for practical live Re-ID implementation in a new context. The COMBINED results are themselves better than COMBINED$_{\text{scaled}}$, which suggest that one should use all the available data to train a good standard Re-ID model for live Re-ID implementation. Finally, we emphasize the good results obtained by training MGN on the COMBINED training dataset. These results are very encouraging after the pessimistic results reported by Sumari et al.\cite{application_level_ReID} for live Re-ID.

\subsection{Impact of the standard Re-ID approaches}

All the approaches tested in this study perform well in the single dataset scenario. However, when it comes to generalization to live operation contexts, MGN has a clear advantage against the other three techniques. This conclusion could already be intuited from the cross-dataset experiments, which suggests a simple yet powerful approach to test future standard Re-ID approaches before live deployment. MGN is the only approach involving a specific image splitting, forcing the network to focus on different body part. In view of our results, this property appears to be desirable for generalization to the live Re-ID setting.

Besides MGN, the SBS approach also appear to present much better generalization than its competitors (Table~\ref{tab:cross_eval} and \ref{tab:live-reid}). Hence, a promising research direction for live Re-ID research would be to design a new standard Re-ID architecture combining features from MGN and SBS, as described in Section~\ref{sec:reid_approaches}.

\section{\uppercase{Conclusion}} \label{sec:conclusion}

\subsection{Overview}
This paper presents a comprehensive evaluation methodology to benchmark different standard Re-ID approaches and training datasets with respect to their ability to be deployed in practical applications from a different context. To do so, we first formalized the new live Re-ID setting, and define new unified evaluation metrics to facilitate interpretation. The performance of different standard Re-ID models is evaluated on this setting. We also conduct simple cross-dataset experiments to see if it can be used to predict which datasets and approaches will generalize better to the live Re-ID setting.

The main conclusions from this study are:
\begin{enumerate}
    \item Although previous work reported very pessimistic results,\cite{application_level_ReID} our experiments showed that it is possible to design good live Re-ID pipelines by properly choosing the standard Re-ID model and combining publicly available training dataset.
    \item Proper choice of the standard re-ID approach and training dataset can influence greatly the results when transferring the model to the cross-domain live Re-ID setting.
    \item Increasing training dataset diversity helps generalization to the cross-domain live Re-ID setting.
    \item Increasing training dataset size allows to improve cross-domain generalization even further.
    \item Simple cross-dataset evaluation can be used to quickly assess the generalization performance of future standard Re-ID techniques for live Re-ID.
\end{enumerate}
Although we only studied the straightforward transfer strategy without fine-tuning, we believe that the results presented here can serve as a good starting point to develop better live Re-ID models in the future.



\subsection{Future work}

The outputs of this study suggest several interesting research directions. First, it would be very valuable to build new live Re-ID datasets, allowing not only to confirm the results obtained in this study, but also to see if good live Re-ID performance is consistent across different scenarios. Then, this benchmark experiment can be extended to account for different pedestrian detectors, another important component of the live Re-ID pipeline. In particular, it would be interesting to study if specific Re-ID approaches combine better with specific object detectors. 

The evaluation methodology proposed in this paper could help to answer this question. 
Another valuable contribution would be to create a ready-to-use website implementing the proposed benchmarking methodology for researchers to test their new approaches easily. 
It would also be interesting to see if existing unsupervised cross-dataset adaptation methods could help generalization to the live Re-ID setting. Finally, it would be interesting to study how the good design choices identified in this study can be leveraged to develop successful end-to-end approaches for live Re-ID.

\bibliographystyle{apalike}
{\footnotesize
\bibliography{example}}



\end{document}